\documentclass[11pt, a4paper, logo, twocolumn, external, copyright]{dm}

\pdftrailerid{redacted}

\makeatletter
\renewcommand\bibentry[1]{\nocite{#1}{\frenchspacing\@nameuse{BR@r@#1\@extra@b@citeb}}}
\makeatother
\usepackage{adjustbox}
\usepackage{multirow}
\usepackage{kantlipsum, lipsum}
\usepackage{dsfont}
\usepackage{gdm-colors}
\usepackage{comment}
\usepackage{wrapfig}
\usepackage[absolute,overlay]{textpos}
\usepackage{amsmath, amssymb}
\usepackage{algorithm}
\usepackage{algorithmic}

\usepackage{wrapfig}
\theoremstyle{definition}

\newtheorem{definition}{Definition}

\usepackage{fancyhdr}
\usepackage{bm}

\definecolor{burgundy}{RGB}{128, 0, 32} % RGB values for burgundy
\definecolor{jhublue}{RGB}{0, 45, 114}

\usepackage[authoryear, sort&compress, round]{natbib}
\hypersetup{
    colorlinks=true,      % Enable colored links
    citecolor=dmblue400,       % Color for citations
    linkcolor=dmblue400,        % Color for internal links
    urlcolor=black       % Color for URLs
}
\graphicspath{{figures/}}

\definecolor{gray}{gray}{.75}
\definecolor{humancolor}{gray}{.95}
\newcommand{\human}[1]{\cellcolor{humancolor}{#1}}

\definecolor{lightblue}{rgb}{0.85,0.9,1}
\colorlet{transparentblue}{lightblue!30}
\newcommand{\gpt}[1]{\cellcolor{transparentblue}{#1}}

\title{GenEx: Generating an Explorable World}

\keywords{\small Generative AI, World Models, Embodied AI, World Explorer}

\author{Taiming Lu}
\author{Tianmin Shu}
\author{Junfei Xiao}
\author{Luoxin Ye}
\author{Jiahao Wang}
\author{Cheng Peng}
\author{Chen Wei}
\author{Daniel Khashabi}
\author{Rama Chellappa}
\author{Alan L. Yuille}
\author{Jieneng Chen}
\affil{Johns Hopkins University}

\vspace{-5cm}
\banner{figures/teaser.pdf}
{
\small{GenEx explores an imaginative world, created from a single RGB image and brought to life as a generated video. See more examples in our \href{https://genex.world/}{\small \textbf{\textcolor{dmred600}{website (genex.world)}}}.}
}
\vspace{-1cm}

\begin{abstract}
{
\vspace{-0.3cm}
\small
{\hskip 2em} Understanding, navigating, and exploring 
the 3D physical \textit{real} world has long been a central challenge in the development of artificial intelligence. In this work, we take a step toward this goal by introducing GenEx, a system capable of 
planning complex embodied world exploration, guided by its \emph{generative imagination} that forms priors (expectations) about the surrounding environments.
{\hskip 2em} GenEx generates an entire 3D-consistent imaginative environment from as little as a single RGB image, bringing it to life through panoramic video streams. Leveraging scalable 3D world data curated from Unreal Engine, our generative model is grounded in the physical world. It captures a continuous 360$^{\circ}$  environment
with little effort
, offering a boundless landscape for AI agents to explore and interact with. GenEx achieves high-quality world generation and robust loop consistency over long trajectories, and demonstrates strong 3D capabilities such as consistency and active 3D mapping.

{\hskip 2em} 
Powered by the generative imagination of the world, GPT-assisted agents are equipped to perform complex embodied tasks, including both goal-agnostic exploration and goal-driven navigation. These agents utilize predictive expectations regarding unseen parts of the physical world to refine their beliefs, simulate different outcomes based on potential decisions, and make more informed choices.

{\hskip 2em} In summary, we demonstrate that GenEx provides a transformative platform for advancing embodied AI in imaginative spaces and brings potential for extending these capabilities to real-world exploration.
}
\end{abstract}

\begin{document}
\maketitle
\newpage
\clearpage
\twocolumn

\section{1. Introduction}
% \begin{figure*}[ht!]
% \begin{center}
% \includegraphics[width=\textwidth]{figures/teaser.pdf}
% \end{center}
% \caption{\small 
% GenEx explores an imaginative world, created from a single RGB image and brought to life as a generated video. See more examples in our \href{https://generative-world-explorer.github.io/}{\small \textbf{\textcolor{dmred600}{website (genex.world)}}}.
% }
% \label{fig:teaser}
% \end{figure*}
Humans explore and interact with the 3D physical world by perceiving their surroundings, taking actions, and engaging with others. Through these interactions, they form mental models that simulate the complexities of their environment. With just a glimpse, humans can construct an internal 3D representation of their surroundings in their minds, enabling reasoning, navigation, and problem-solving. This remarkable ability has long been a central challenge in the development of artificial intelligence.  

In this work, we introduce \textbf{GenEx}, a platform designed to push this boundary by \textbf{Gen}erating an \textbf{Ex}plorable world and facilitating explorations in this generated world. GenEx combines two interconnected components: an imaginative world, which dynamically generates 3D environments for exploration, and an embodied agent, which interacts with this environment to refine its understanding and decision-making. Together, these components form a symbiotic system that enables AI to simulate, explore, and learn in ways similar to human cognitive processes.

We begin by constructing an imaginative world that captures a 360$^{\circ}$, 3D environment grounded in the physical world, leveraging recent advancements in Generative AI. Starting from a single image, the model generates new environments expansively and dynamically while maintaining coherence and 3D consistency, even during long-distance exploration. This boundless landscape provides endless opportunities for AI agents to explore and interact.

The environment is brought into life in the form of diffusion video generation, conditioned on moving angle, distance, and a single initial view to serve as a starting point. To address field-of-view constraints, we utilize panoramic representations and train our video diffusion models with spherical-consistent learning techniques. This ensures the generated environments maintain coherence and 3D consistency, even during long-distance exploration. To anchor our video generation model in the physical world, we curate training data from physics engines like Unreal Engine, enabling realistic and immersive outputs.

Within this imaginative landscape, embodied agents play a crucial role. Enhanced by GPTs, these agents can explore unseen parts of the physical world with imagined observations to refine their understanding of surroundings, simulate different outcomes based on potential decisions, and make more informed choices. Furthermore, GenEx supports multi-agent scenarios, allowing agents to mentally navigate others' positions, share imagined beliefs, and collaboratively refine their strategies.

In summary, GenEx represents a transformative step forward in the development of AI, offering a platform that bridges the generative and physically grounded world. By enabling AI to explore, learn, and interact in boundless, dynamically generated environments, GenEx opens the door to applications ranging from real-world navigation, interactive gaming, and VR/AR to embodied AI.

\section{2. Generating an Explorable World}
We define the explorable generative world and the problem in $\S$~\ref{sec:GenEx_formulation}, present the world initialization in $\S$~\ref{sec:single-to-pano} and world transition in $\S$~\ref{sec:world_transition}.

\subsection{Problem Formulation} 
\label{sec:GenEx_formulation}
\textbf{Defining an explorable generative world}.  
We define an explorable generative world as an  AI-generated virtual environment, constrained to the agent’s immediate surroundings. The generative world is both physically plausible and visually coherent. This environment is represented by the agent’s egocentric panoramic observations, denoted as $\mathbf{x}$. 
While $\mathbf{x}$ is synthesized, it remains grounded in intuitive physical principles and realistic appearance, akin to a high-fidelity, physically realistic video game environment. 

Crucially, the explorable nature of our generative world ensures the agent’s experience is not limited to a static scene. Instead, the environment dynamically evolves in response to the agent’s movements and actions, simulating continuous and coherent exploration. 
Formally, let $a_t$ be the agent’s action at step $t$, encompassing both view rotation $\alpha$ and forward distance $d$. Let $\mathbf{x}_t$\,$=$\,$ (x_t^0, x_t^1, \dots, x_t^S)$ represent the sequence of panoramic observations encountered as the agent moves according to $a_t$, where $S$ corresponds to sequence length in $\mathbf{x}_t$, or the traveled distance. Each $x_t^s$ in $\mathbf{x}_t$ is generated to reflect the environment’s currently perceivable state, ensuring that the agent’s evolving viewpoint remains coherent and physically meaningful. 

We train our models using data harvested from a controlled, simulated setting. By employing a physics-based data engine (\S\ref{sec:data_engine}), we ensure realistic and diverse training scenarios that capture the intricate variations encountered in complex, virtual landscapes.

\textbf{Task formulation}: We reformulate the task of ``exploring a generative world'' as the problem of generating an initial panoramic world view $x_0$ and a sequence of world views represented by panoramic videos $\textbf{x}_{1:T}$, together represented as $\mathbf{x}_{0:T}$, given a single initial image $i_0$, a description $l_0$, and action $a_t$ at each step $t$, where $t=1,\ldots,T$. 
Formally, we have
\vspace{5mm}
\begin{align*}
p(\textbf{x}_{0:T} \mid i_0, l_0) 
= \underbrace{p_{\theta_1}(x_0 \mid i_0, l_0)}_{\text{world initialization}} \underbrace{\prod_{t=1}^{T} p_{\theta_2}\bigl(\textbf{x}_t \mid x_{t-1}^{S}, a_t\bigr)}_{\text{world transition}}.
\end{align*}
\vspace{5mm}

In this unified form, the core terms are:
\begin{itemize}[leftmargin=*,itemsep=0pt, before=\vspace{-0.6\baselineskip}, after=\vspace{-\baselineskip}]
\item \textbf{World initialization} (\S\ref{sec:single-to-pano}): 
Given the initial image $i_0$ and a language description $l_0$, the anchor 360$^\circ$ world view $x_0$ is sampled from:
\[
x_0 \sim p_{\theta_1}(x \mid i_0, l_0),
\]
where $\theta_1$ is an image-to-panorama generator. 

\vspace{0.3cm}
\item \textbf{World transition} (\S\ref{sec:world_transition}): 
Given the chosen action $a_t$, the next world view $\mathbf{x}_t$ is sampled from:
\[
\mathbf{x}_t = (x_t^0, x_t^1, \dots, x_t^{S}) \sim p_{\theta_2}(\textbf{x} \mid x_{t-1}^{S}, a_t),
\]
where $\theta_2$ is a 360$^\circ$ panoramic video generator, $t=1,\ldots, T$, and $x_0^S \coloneq x_0$.
\end{itemize}

\begin{algorithm}[ht!]
\small
\caption{Generating an Explorable World $p(\textbf{x}_{0:T} \mid i_0, l_0)$}
\label{alg:exploration}
\begin{algorithmic}[1]
\REQUIRE
\begin{itemize}[leftmargin=*,itemsep=0pt]
 \item A initial single-view image $i_0$. \\
 \item A language description $l_0$ specifying the desired panoramic world initialization. \\
 \item A conditional distribution $p_{\theta_1}(x \mid i_0,l_0)$, parameterized by an image-to-panorama generation model $\theta_1$ to initialize the 360$^\circ$ world. \\
  \item   Action space $\mathcal{A}$ defined in the physical engine, from which an action is sampled: $a_{t} \sim \mathcal{A}$.
 \item A conditional distribution $p_{\theta_2}(\mathbf{x} \mid x_{t-1}^{S},a_t)$, parameterized by a panoramic video generation model $\theta_2$. 
\end{itemize}

\STATE Notation: Let $\mathbf{x}_t = (x_t^0, x_t^1, \dots, x_t^{S})$ denote the generated panoramic video at exploration step $t$.  
Here, $x_t^{S}$ is the latest explored panoramic view.

\STATE \textbf{World initialization}: Initialize a 360$^\circ$ panoramic world from a single image:
$$x_0 \sim p_{\theta_1}(x \mid i_0, l_0)$$
\FOR{$t = 1$ to $T$}
    \STATE \textbf{World transition} at step $t$: Given $a_t \sim \mathcal{A}$ and the latest explored world $x_{t-1}^{S}$ where $x_0^S \coloneq x_0$, we sample the new panoramic video $\mathbf{x}_t$:
$$
\mathbf{x}_t \sim p_{\theta_2}(\mathbf{x} \mid x_{t-1}^{S}, a_t)
$$

\ENDFOR
\RETURN The initial 360$^\circ$ panoramic world view $x_0$ and a sequence of generated panoramic states $\mathbf{x}_{1:T}$, which together represent one explorable generative world, denoted as $\mathbf{x}_{0:T}$.
\end{algorithmic}
\end{algorithm}

\vfill\eject
\subsection{World Initialization}  
\label{sec:data_engine}
\begin{figure}[ht!]
\begin{center}
\vspace{-3mm}
\includegraphics[width=\linewidth]{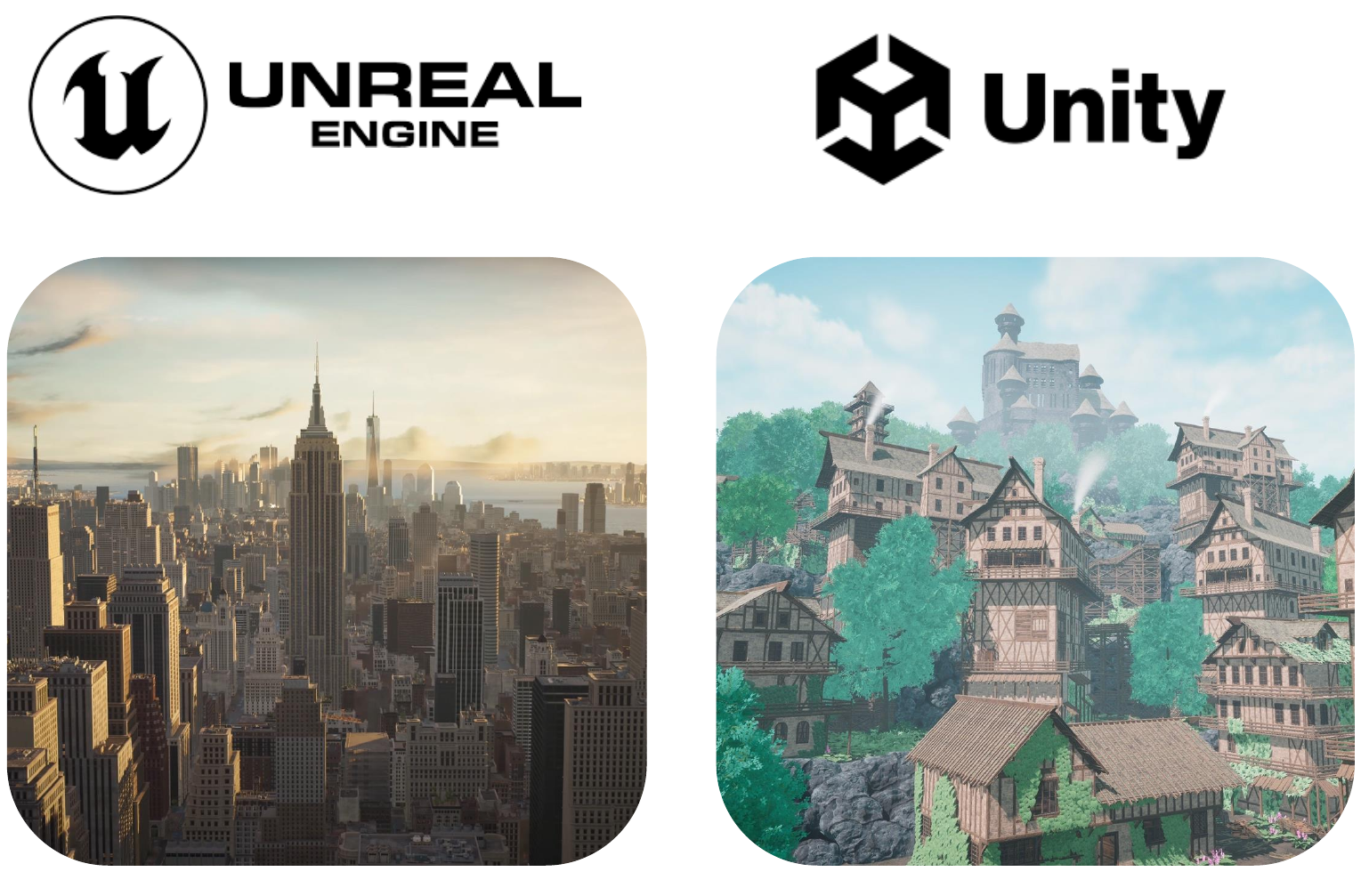}
\end{center}
\vspace{-8mm}
\caption{\small Our data curation leverages physical engines, utilizing realistic city assets from UE5 and animated world assets from Unity.
}
\label{fig:data_engine}
\end{figure}

\textbf{Preliminary: data and representation}.
Collecting diverse world exploration data in the real world is challenging due to resource constraints and environmental variability. Thus, we utilize physics engines such as Unreal Engine 5 and Unity in \autoref{fig:data_engine} for data curation. These engines allow for the creation of rich, diverse virtual environments where we can simulate exploration trajectories and collect corresponding data efficiently.

We represent the 360$^\circ$ world using the panoramic view of the agent. Panoramic images capture a complete 360$^\circ$ $\times$ 180$^\circ$ view of a scene from a fixed viewpoint. One common panoramic representation is the \textit{cubemap}, which projects a 360$^\circ$ view onto the six faces of a cube. Each face captures a 90$^\circ$ field of view, resulting in six perspective images that can be seamlessly stitched together. Due to its simplicity and compatibility with rendering engines, we directly collect cubemaps in the physics engine to represent the egocentric world. Notably, cubemaps, equirectangular panorama, and sphere are three representations of 360$^\circ$ 
 panoramic world. The curated cubemaps will be projected to equirectangular panoramas for video generation in the world exploration stage, and projected to spherical space when changing the exploring angle.

\begin{figure}[b!]
\begin{center}
\includegraphics[width=\linewidth]{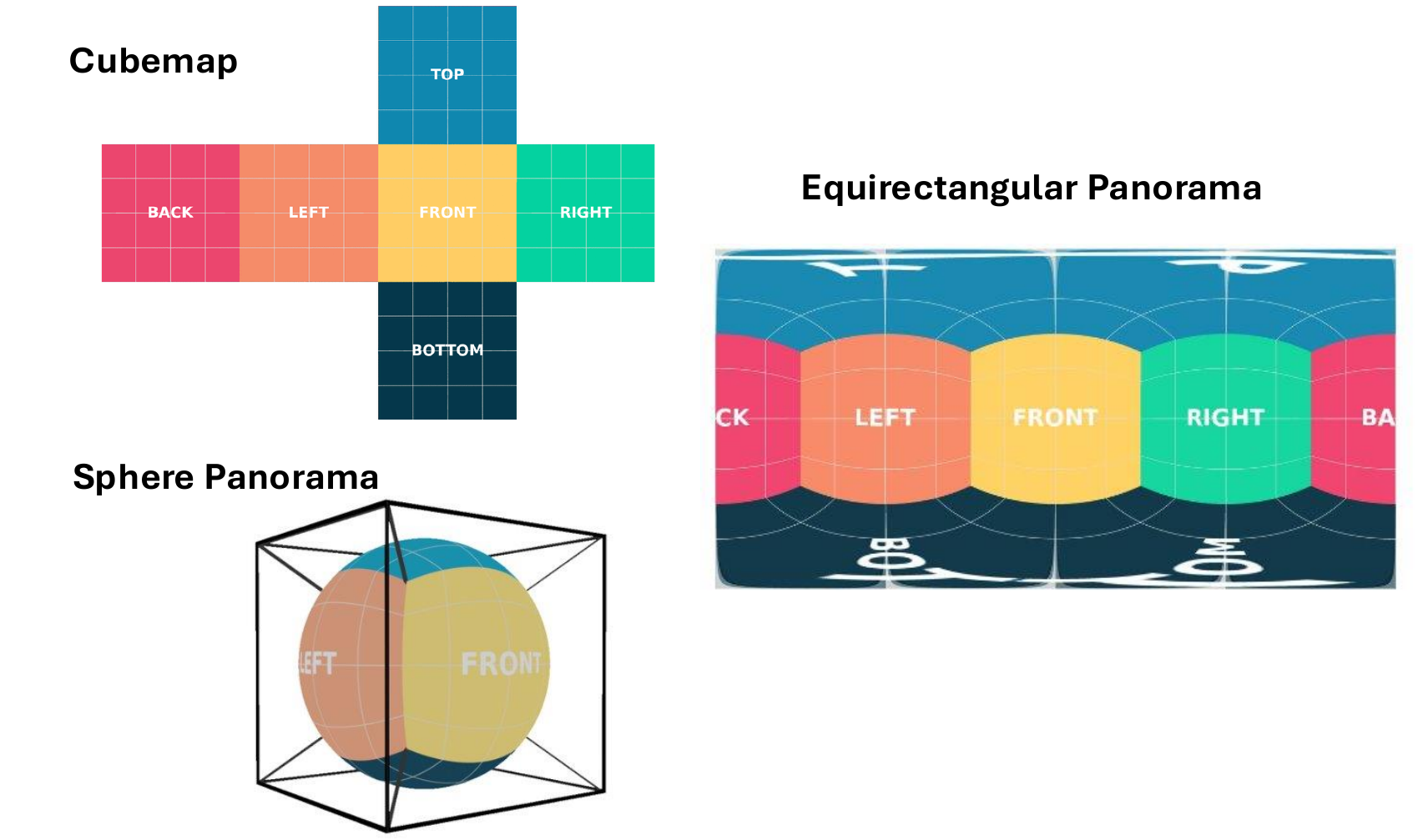}
\end{center}
\vspace{-5mm}
\caption{\small Three panorama representations that can be transformed into one another.
}
\label{fig:panorama_demo}
\end{figure}

Given predefined exploration trajectories, we collect sequences of cubemaps to represent different exploration outcomes in the virtual world. By sampling a large number of exploration directions uniformly, we curate an extensive dataset of world exploration scenarios, which serves as the training data for our models.

\textbf{World initialization model}.
Starting from a single input image $i_0$, we aim to construct a full 360$^\circ$ panoramic representation $x_0$ of the agent’s environment. To achieve this, we condition a pretrained text-to-image diffusion model on both the input image $i_0$ and a text description $l_0$ of the desired 3D world, yielding a high-dynamic-range panorama. Thus, $x_0$ is drawn from the conditional distribution $p(x \mid i_0, l_0)$. 

\label{sec:single-to-pano}
\begin{figure}[ht!]
    \centering
    \includegraphics[width=\linewidth]{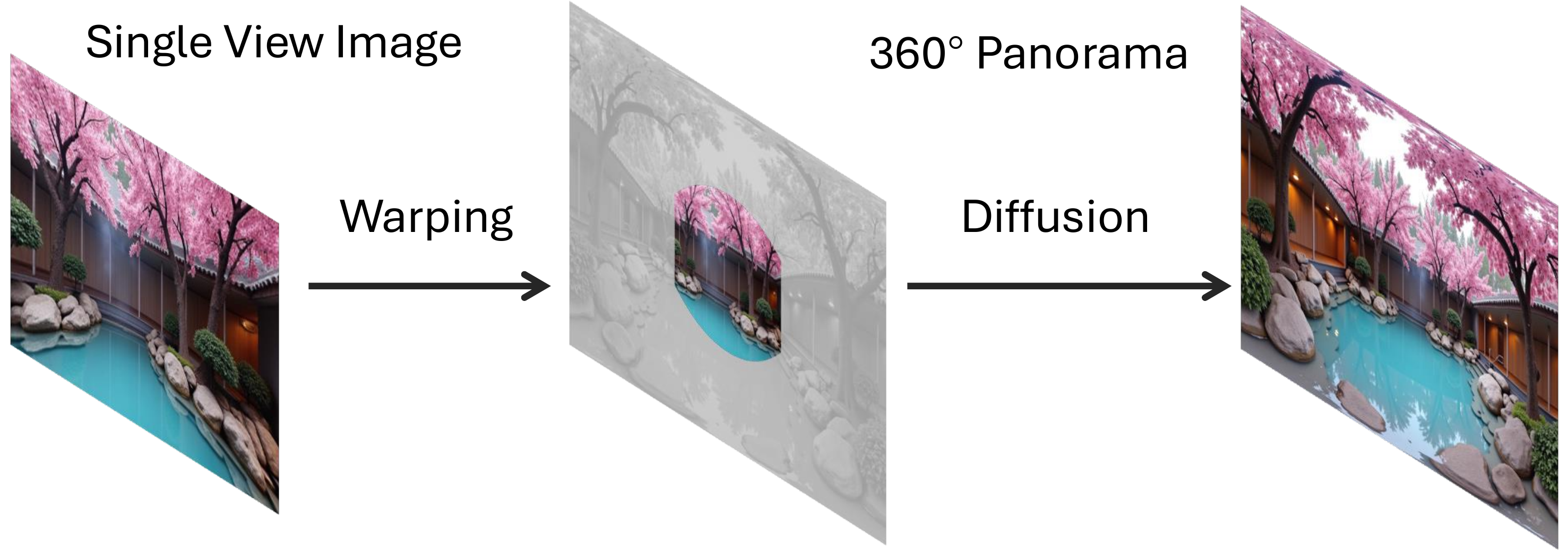}
    \caption{\small From single view to 360$^\circ$ panorama.}
    \label{fig:enter-label}
\end{figure}

Our world initialization model is built up on a state-of-the-art text-to-panorama model ~\citep{bilcke2024flux1devpanoramalora} tuned from the state-of-the-art text-to-image model FLUX.1~\citep{blackforestlabs2024flux1dev}. The text-to-panorama model~\citep{bilcke2024flux1devpanoramalora} generates a panorama from a text description $l_0$: 
$$x_0 \sim p_{\text{flux}}(x \mid l_0)\,.$$
However, without being conditioned on the single image, this approach cannot guarantee the coherence of generated panorama $x_0$ with the provided reference image $i_0$. 

We extend the model to condition on both textual input and a single image. This adaptation allows the model to produce a full 360-degree environment that aligns with the provided image:
$$
    x_0 \sim p_{\theta_1}(x \mid i_0, l_0)\,.
$$
Although this yields a coherent, image-consistent panorama, the scene remains static and does not permit dynamic movement or exploration. To enable deeper interaction within the generative world, we introduce the world transition.

\clearpage
\subsection{World Transition}
\label{sec:world_transition}

When the agent moves within the imaginative environment, its egocentric 360$^{\circ}$ view changes, prompting a \textit{world transition.} We model this transition as an action-driven panoramic video generation process, transforming the previously observed panorama into a new, forward-looking view as the agent progresses.

\textbf{Transition objective}. The goal is to sample $\mathbf{x}_t =(x_t^0, x_t^1, ..., x_t^{S})$, a newly explored panoramic video, conditioned on the previous panorama $x_{t-1}^{S}$ and the action $a_t=(\alpha_t, d_t)$. Here, $\alpha_t$ is the moving angle and $d_t$ is the distance.
Formally, we have the transition objective:
$$
\mathbf{x}_t \sim p(\mathbf{x} \mid x_{t-1}^S, a_t)\,.
$$ 
The transition procedure has core modules:
\begin{itemize}[leftmargin=*,itemsep=0pt, before=\vspace{-0.6\baselineskip}, after=\vspace{-\baselineskip}]
\item \textbf{Action sampling}: Consider an action sequence $a_{1:T}$ drawn from an infinitely large action set in the Unreal Engine and Unity. We can denote the action space as:
$\mathcal{A}$, where $|\mathcal{A}| = \infty$.
Each element of the sequence for \( t = 1,\ldots,T \) is sampled from $\mathcal{A}$:

\[
a_t \sim \mathcal{A}, \quad t=1,\ldots,T,
\]
As a result, the entire action sequence $a_{1:T}=(a_1, \ldots, a_T)$ lies in $\mathcal{A}^{T}$.

\item \textbf{Sphere rotation}: The action $a_t$ 
  determines a rotation angle $\alpha_t$, which we apply to the spherical representation of the equirectangular panorama $x_{t-1}^{S}$. This yields a rotated equirectangular panorama ${x_{t-1}^{S}}^\prime$: 
$${x_{t-1}^{S}}^\prime =\mathcal{T}(x_{t-1}^{S}, \alpha_t)\,,$$ where $\mathcal{T}$ is a known rotation geometric transformation defined to \autoref{eq:composite_rotation_T_appendix} in Appendix.
\item \textbf{Panoramic video generation}: We next generate videos to travel in the imaginative space by distance $d_t$. Our video generator is adapted from a video diffusion model conditioned on the latest explored view ${x_{t-1}^{S}}^\prime$ and randomly sampled noise $\epsilon_t \sim \mathcal{N}(0, I)$:
$$\mathbf{x}_t \sim p_{\theta_2}(\mathbf{x} \mid {x_{t-1}^{S}}^\prime, \epsilon_t)\,.$$
This approach ensures that each generated panoramic video remains consistent with the prior view, while incorporating stochastic variations to represent an explorable world.
\end{itemize}

\begin{figure}[ht!]
\begin{center}
\vspace{0.7cm}
\includegraphics[width=0.98\linewidth]{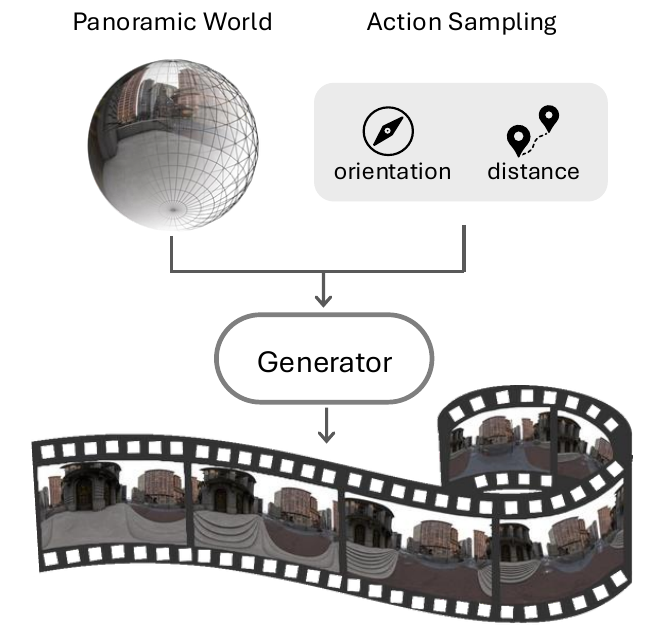}
\end{center}
\caption{\small We model the world transition as a panoramic video generation process. Given the last explored 360$^{\circ}$ panorama and an action that rotates the viewing sphere, the model produces a sequence of newly generated panoramic views
}
\label{fig:pano_gen_infer}
\end{figure}

We aim to learn to produce panoramic videos that remain visually coherent on a spherical surface. Without additional constraints, training on equirectangular panorama alone can result in discontinuities at the panorama edges. To address this, we adopt spherical-consistency learning, or SCL, detailed in \citep{lu2024generative}, which promotes smooth and continuous imagery across all viewing directions on the sphere.

\paragraph{Summary.}
In essence, the world transition step updates the agent’s observed 360° panorama into a newly explored view sequence. Through action-driven rotation, spherical adjustments, and a diffusion-based video model, we achieve seamless transitions and maintain coherent, panoramic representations as the agent navigates the generative environment.

\clearpage
\section{3. Exploration in the Generative World}
\label{sec:exploration}
After generating the explorable world, human or embodied agents can explore the virtual world with an exploration policy, defined in \S\ref{sec:explore_policy}. We then introduce three exploration modes in \S\ref{sec:explore_mode}.
\subsection{Exploration Policy}
\label{sec:explore_policy}

% \chen{missing some starting words}
The exploration action $a_t$ is decided by a policy: 
$$a_t = \arg\max_a \pi_{explore}(a|x_{t-1}^{S}, \mathcal{I}),$$ 
where $\mathcal{I}$ is the instruction that specifies the exploration mode to be either human interaction or assisted by a GPT, detailed in \S\ref{sec:explore_mode}. Note that $x_{t-1}^{S}$ denotes the latest explored view from the previous step $t-1$. At $t=1$, it corresponds to the initial panorama $x_0$. The action $a_t = (\alpha_t, d_t)$ defines how the agent rotates its field of view with the rotation angle $\alpha_t$ and moves forward with $d_t$ distance, shaping the direction and extent of exploration.

\subsection{Exploration Modes}
\label{sec:explore_mode}
The GenEx framework enables agents to explore within an imaginative world by streaming video generation, based on current single view image $i_0$ and the given exploration action $a$.  
% $$
% \textbf{x} \sim p_{GenEx}(\textbf{x} \mid i_0, g_{nav}).
% $$

\begin{figure}[ht!]
\begin{center}
\includegraphics[width=\linewidth]{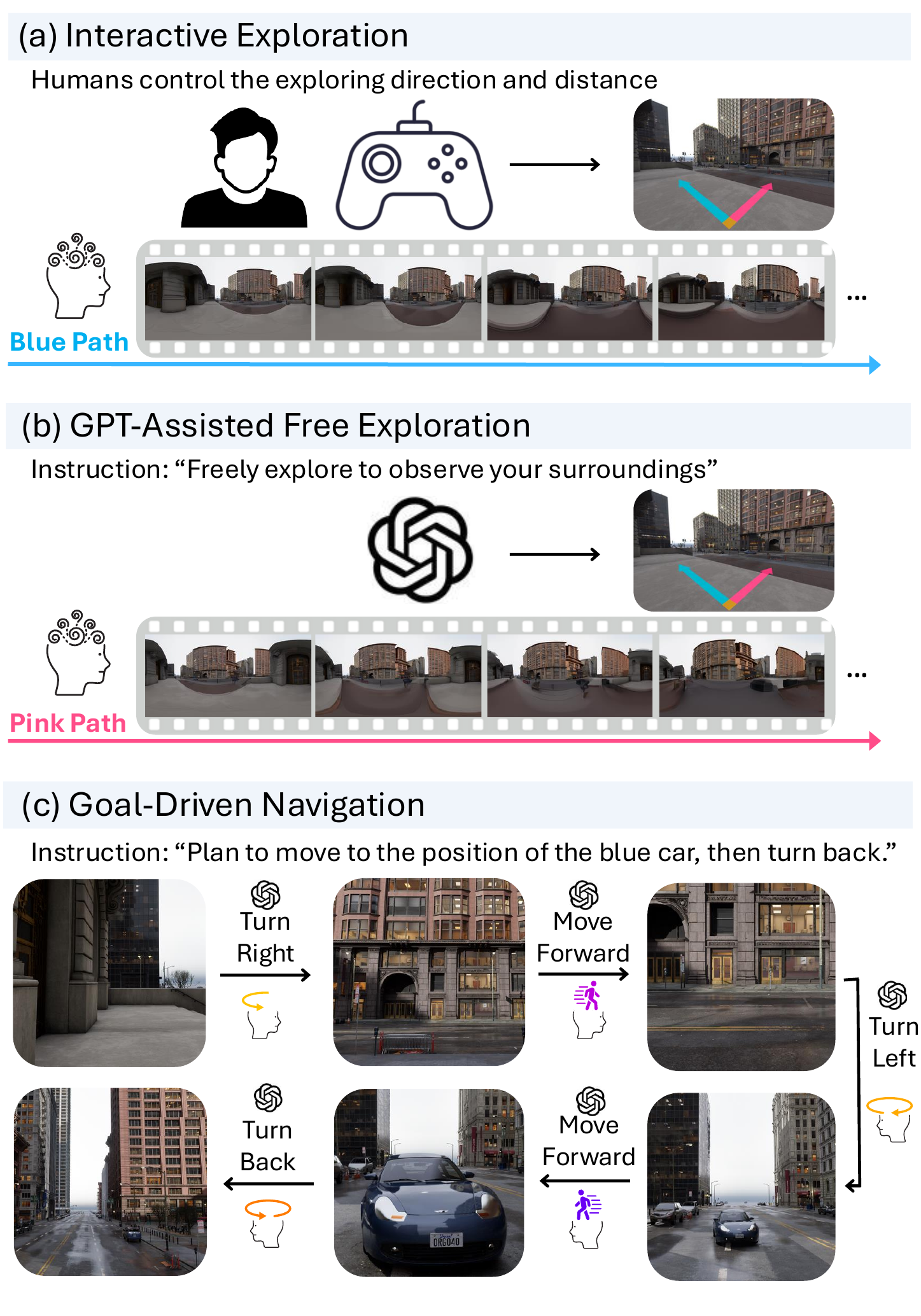}
\end{center}
\caption{\small Three exploration modes --- interactive, GPT-assisted, and goal-driven --- each defined by distinct exploration instructions.
% \chen{case c seems not clearer enough. Adding more explanations may help? }
}
\label{fig:explore_mode}
% \vspace{-0.3cm}
\end{figure}

We support three modes for generative world exploration, including (a) interactive exploration, (b) GPT-assisted free exploration, and (c) goal-driven navigation, illustrated in \autoref{fig:explore_mode}. 

\textbf{Interactive exploration.} % goal-agnostic
GenEx enables the agent to freely explore the synthetic world with an unlimited range of orientations, enhancing its understanding of the surrounding environment. Users can control the agent's movement directions and distances, allowing for continuous exploration of the virtual world.

\textbf{GPT-assisted free exploration.}
However, human-provided commands can sometimes lead the model to collapse. For example, if users instruct the agent to move excessively close to a wall, the resulting viewpoint may reduce the quality of subsequent generated video frames.

To mitigate this, we employ a GPT-4o~\citep{achiam2023gpt} as a ``pilot'' to determine exploration configurations, encompassing full 360$^\circ$ explorable directions and distances. Given that generation quality can compoundingly degrade over time, GPT-4o acts as a policy that selects actions to maximize the fidelity of generative worlds and avoid model collapsing.

\textbf{Goal-driven navigation.} % goal-driven
The agent receives a goal with navigation instruction $\mathcal{I}$, such as, ``Move to the blue car's position and orientation." 
GPT performs high-level planning based on the instruction and initial image, generating low-level exploration configurations in an iterative manner. GenEx then processes these configurations step-by-step, updating images progressively throughout the imaginative exploration. This allows for greater control and targeted exploration.

\clearpage
\section{4. Advancing Embodied AI}
\label{sec:eai}

In our generative world, we can explore previously unobserved regions of the physical environment, gather more comprehensive information, and refine our beliefs for more informed decision-making. We frame this process in a form of human-like decision-making—an ``imagination-augmented policy''—that could play a crucial role in shaping the future of embodied AI.

\textbf{Preliminary}. We first denote a common embodied policy as $\pi_{\theta}(A|o, g)$ where $\theta$ is a GPT-based planner, $o$ is the agent’s observation, $g$ is the goal to answer questions such as ``Danger ahead. Stop or go ahead?''. Here, $A$ denotes higher-level embodied actions (\textit{e.g.,} answering the questions or generating navigation plans), which differ from the exploration actions $a$ introduced earlier. However, if the observation is limited to a single initial image $i_0$, then executing $\arg\max_A \pi_{\theta}(A|o=i_0, g)$ may fail because it provides no visibility into unseen parts of the environment.

\begin{figure}[ht!]
\centering
% \hfill
\includegraphics[width=1.1\linewidth]{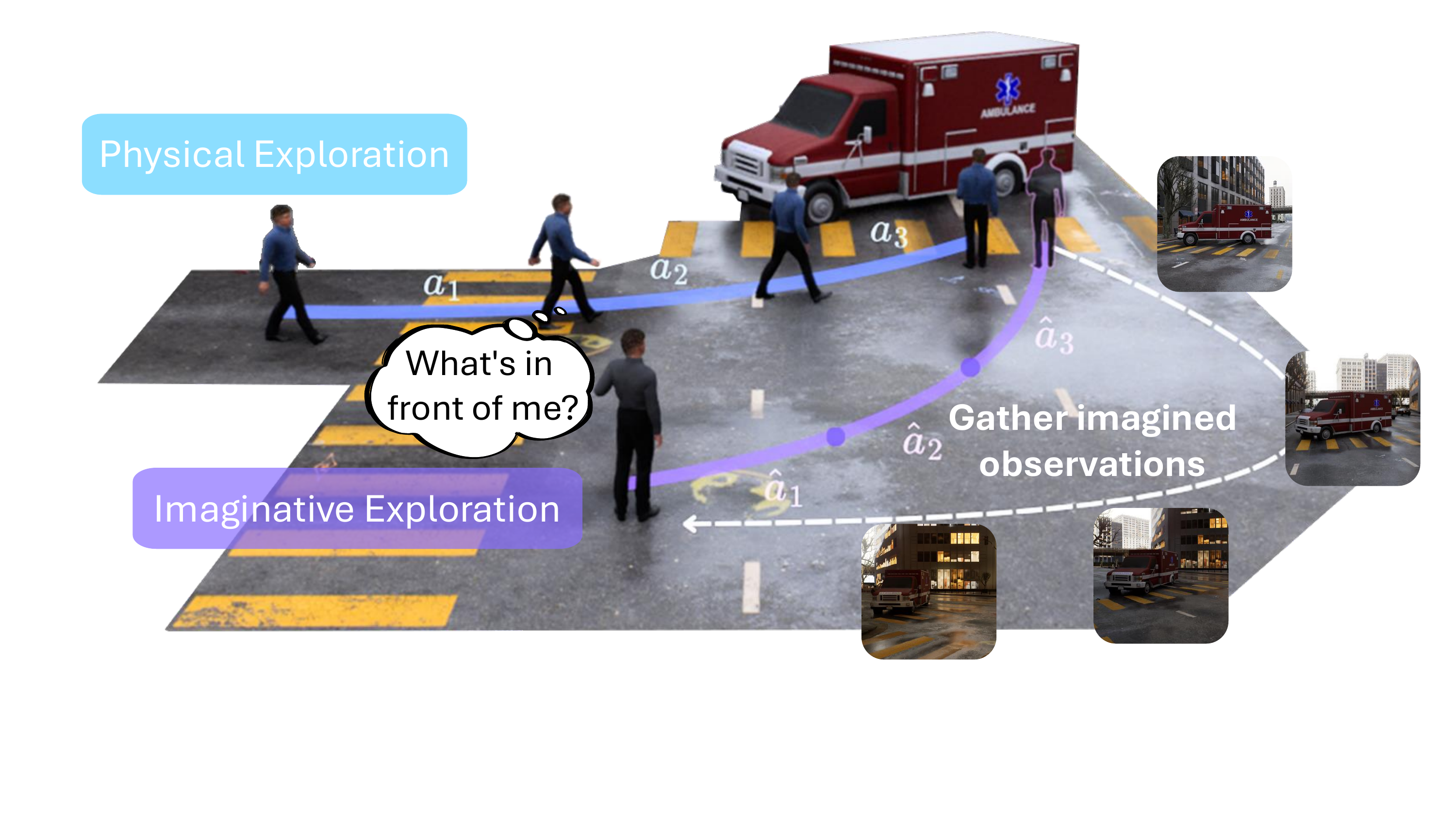}
\vspace{-15mm}
\caption{GenEx-driven imaginative exploration can gather observations that are just as informed as those obtained through physical exploration.}
\label{fig:imagination_state}
\end{figure}

The decision can become more informed if the agent gains a clearer understanding of its surroundings~\citep{fan2024evidential}. By navigating through the physical space, the agent gathers additional information about its environment (\textcolor[RGB]{130,151,246}{ ``Physical'' path in the cyan color} in  
 \autoref{fig:imagination_state}), enabling more accurate assessments and better choices moving forward. 

Nevertheless, physically traversing the space is inefficient, expensive, and even impossible in dangerous scenarios. To streamline this process, we use imagination as a pathway for the agent to simulate outcomes without physically traversing (\textcolor[RGB]{174,153,246}{ ``Imaginative'' path in purple color} in \autoref{fig:imagination_state}).

\indent The key question is: 
\noindent\vspace{-0.3cm}
\begin{center}
\textit{How can an agent make more informed decisions through \textbf{exploration} in a generative 360$^\circ$ world?}
\end{center}
\vspace{-0.5cm}

\begin{algorithm}[ht!]
\small
\caption{Imagination-Augmented Policy}
\label{alg:imaginative_policy}
\begin{algorithmic}[1]
\REQUIRE 
\begin{itemize}[leftmargin=*,itemsep=0pt]
 \item Initial observation $i_0$ and world initialization description $l_0$
 \item A goal $g$ to answer embodied questions. \textcolor{lightgray}{\emph{E.g}, ``Danger ahead—stop or go ahead?''}
 \item A navigation instruction $\mathcal{I}$. \textcolor{lightgray}{\emph{E.g}, ``Navigate to the unseen parts of the environment.''}
\item GenEx $p(\textbf{x}_{0:T} | i_0, l_0, \mathcal{I})$ defined in $\S~\ref{sec:GenEx_formulation}$ and \autoref{alg:exploration}.
 \item An embodied policy $\pi_{\theta_3}(A|o, g)$ conditioned on observation variable $o$ and goal $g$.
\end{itemize}

\STATE \textbf{Gather imagined observations} with GenEx:
\[
\textbf{x}_{0:T} \sim p(\textbf{x}_{0:T} \mid i_0, l_0, \mathcal{I})
\]

\STATE \textbf{Select an action with imagined observations} to maximize the policy:
\[
A = \arg\max_A \pi_{\theta}(A \mid i_0, \textbf{x}_{0:T}, g)
\]
\vspace{-0.3cm}
\end{algorithmic}
\end{algorithm}

\subsection{Imagination-Augmented Policy}
\label{sec:imaginative_policy}
\vspace{-0.1cm}
We propose a new policy based on imagined observations in the generative world, described in \autoref{alg:imaginative_policy}.
The Imagination-Augmented Policy consists of the following two steps:
\begin{itemize}[leftmargin=*,itemsep=0pt, before=\vspace{-0.6\baselineskip}, after=\vspace{-\baselineskip}]
\item \textbf{Step 1}: Gather imagined observations sampled
from GenEx (\autoref{alg:exploration}):
\[
\textbf{x}_{0:T} \sim p(\textbf{x}_{0:T} \mid i_0, l_0, \mathcal{I})\,.
\]
\item \textbf{Step 2}: Select an action conditioned on the imagined observations to maximize the policy:
\[
A = \arg\max_A \pi_{\theta_3}(A \mid i_0, \textbf{x}_{0:T}, g)\,.
\]

\end{itemize}

\begin{figure*}[ht!]
\begin{center}
\includegraphics[width=0.99\linewidth]{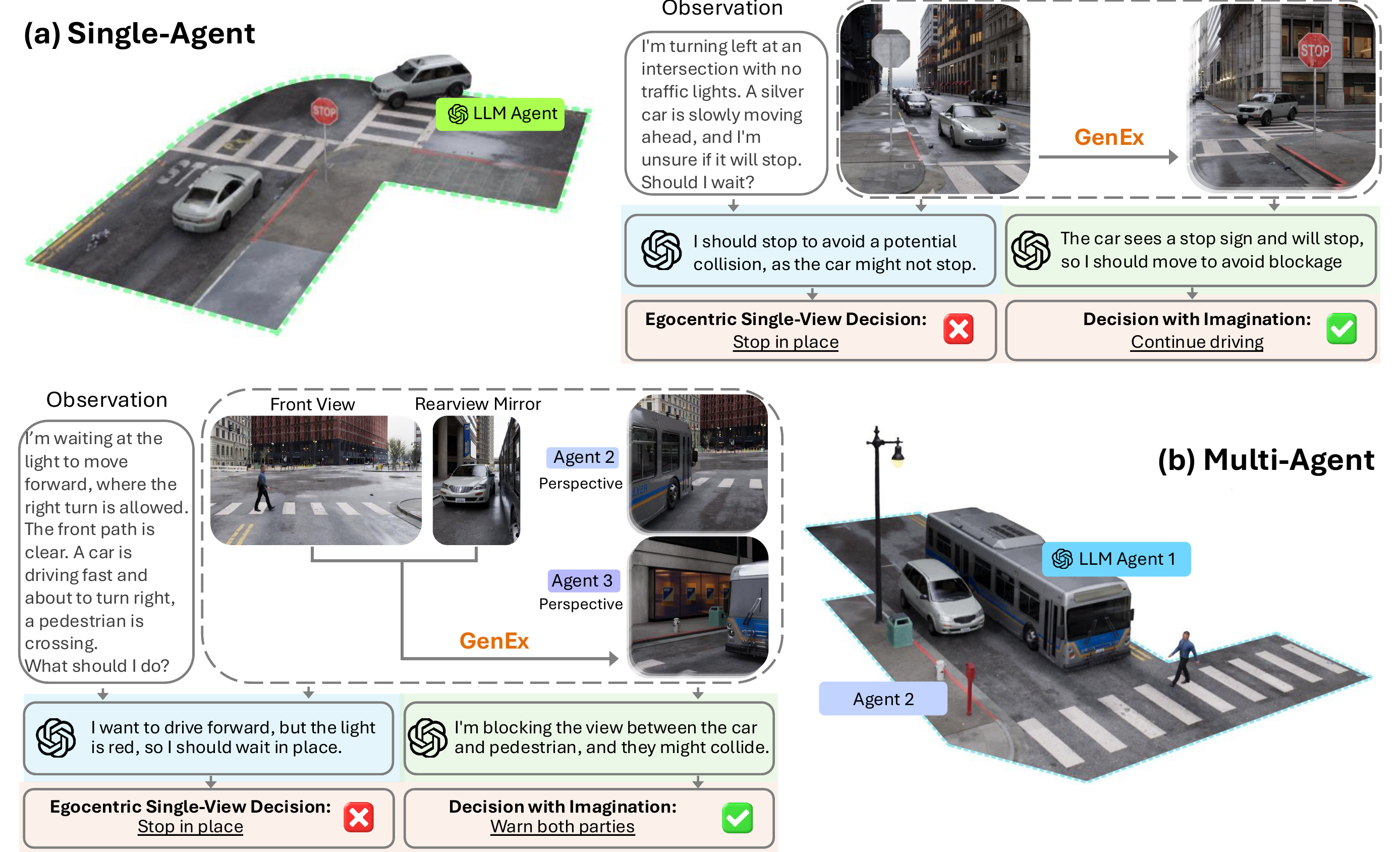}
\end{center}
\vspace{-5mm}
\caption{Single agent reasoning with imagination and multi-agent reasoning and planning with imagination. (a) The single agent can imagine previously unobserved views to better understand the environment. (b) In the multi-agent scenario, the agent infers the perspective of others to make decisions based on a more complete understanding of the situation. Input and generated images are panoramic; cubes are extracted for visualization.}
\label{fig:embodied_agent_examples}
\end{figure*}

In our work, we apply GenEx for imaginative exploration and an LMM as the policy model $\pi_{\theta_3}$, with examples in \autoref{fig:embodied_agent_examples}.

Compared to $\arg\max_a \pi_{\theta_3}(A \, |\, i_0, g)$ the common policy which selects the action based solely on real observations $i_0$, the Imagination-Augmented Policy selects actions using both actual and imagined observations $(i_0, \textbf{x}_{0:T})$, potentially leading to more informed decisions.
\vspace{-5mm}

\subsection{\small Multi-Agent Imagination-Augmented Policy}
\label{sec:multiagent_policy}
Our Imagination-Augmented Policy can be generalized to the multi-agent scenario. An agent can explore the position of other agents. This predicts other agents' observations and infers their understanding of the surrounding environments.

Technically, we can create multiple exploration paths by providing instructions like ``navigate to the position of agent-k''. The agent can then explore the generated $360^\circ$ environment to reach agent-k's location.

By extending \autoref{alg:imaginative_policy}, the Multi-Agent Imagination-Augmented Policy has three steps:

\begin{itemize}[leftmargin=*,itemsep=0pt, before=\vspace{-0.6\baselineskip}, after=\vspace{-\baselineskip}]
\item \textbf{Step 1}: Gather imagined observations by exploring the position to agent-k using \autoref{alg:exploration}, with instruction $\mathcal{I}_k$ ``navigate to the position of agent-k'':
\[
\textbf{x}_{0:T}^{(k)} \sim p(\textbf{x}_{0:T} \mid i_0, l_0, \mathcal{I}_k)\,.
\]

\item \textbf{Step 2}: Repeat Step 1 a total of $K$ times, then imaginatively explore the resulting positions of all $K$ agents in our generated explorable world:

\[
\{\textbf{x}_{1:T}^{(k)}\}_{k=1}^K = (\textbf{x}_{1:T}^{(1)}, \textbf{x}_{1:T}^{(2)},...,\textbf{x}_{1:T}^{(K)})\,.
\]
\vspace{1mm}

\item \textbf{Step 3}: Select an embodied action $A$ with imagined observations to maximize the policy:
\[
A = \arg\max_A \pi_{\theta_3}(A \mid i_0, \{\textbf{x}_{1:T}^{(k)}\}_{k=1}^K, g)\,.
\]
\end{itemize}

When exploring another agent's surrounding environment, we can predict what that agent sees, understands, and might do next, which in turn helps us adjust our own actions with more complete information.

\clearpage

\section{5. Applications}
\subsection{Generation Quality}
We evaluate the \textbf{video generation quality} using FVD~\citep{unterthiner2019accurategenerativemodelsvideo}, SSIM~\citep{SSIM}, LPIPS~\citep{zhang2018unreasonableeffectivenessdeepfeatures}, and PSNR~\citep{PSNR}. \autoref{tab:video_quality_metrics} shows our earlier GenEx version~\citep{lu2024generative} has high video quality in all metrics. 

\begin{table}[ht!]
\centering
\renewcommand{\arraystretch}{1.5}
\setlength{\tabcolsep}{3pt}
\resizebox{0.49\textwidth}{!}{
\begin{tabular}{lcccccc}
    \toprule
    Model & Representation & FVD $\downarrow$ & MSE $\downarrow$ & LPIPS $\downarrow$ & PSNR $\uparrow$ & SSIM $\uparrow$   \\
    \midrule
    Baseline & 6-view cubemaps & 196.7 & 0.10 & 0.09 & 26.1 & 0.88  \\ % checked
    GenEx w/o SCL & panorama &  81.9 & 0.05 & 0.05 & 29.4 & 0.91 \\
    \textbf{GenEx} & panorama & \textbf{69.5} & \textbf{0.04} & \textbf{0.03} & \textbf{30.2} & \textbf{0.94}  \\ % checked
    \bottomrule
\end{tabular}
}
\caption{\small GenEx with high generation quality.}
\label{tab:video_quality_metrics}
\vspace{-4mm}
\end{table}

\subsection{Exploration Loop Consistency}
\begin{figure}[ht!]
    \centering
    \includegraphics[width=0.4\textwidth]{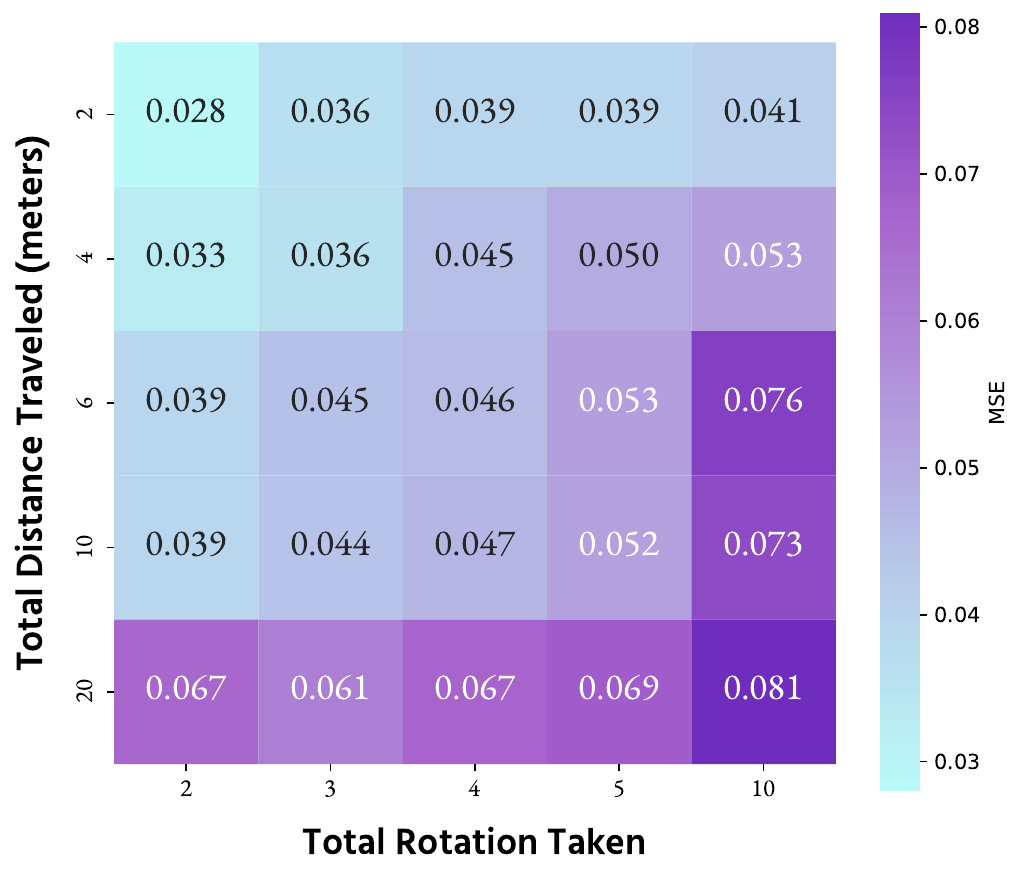}
    \caption{\small  Imaginative Exploration Loop Consistency (IELC) varying distance and rotations.}
    \label{tab:cycle_consistency_mse}
\end{figure}

% \textbf{Loop closure consistency}.  
We propose \textbf{Imaginative Exploration Loop Consistency} (IELC) to measure long-range exploration fidelity. For each randomly sampled closed-loop path, we compute the latent MSE between the initial real image and the final generated image, and then average these values over 1000 loops with varying rotations and distances, discarding blocked paths. As shown in \autoref{tab:cycle_consistency_mse}, the IELC remains high even for 20m loops and multiple consecutive videos, maintaining latent MSE below 0.1 and thus indicating minimal drift. This robustness stems from preserving spherical consistency, ensuring that rotations do not compromise image quality.

\subsection{Generating Bird's-Eye Worlds}
\begin{figure}[h]
\centering
\vspace{-3mm}
\includegraphics[width=1\linewidth]{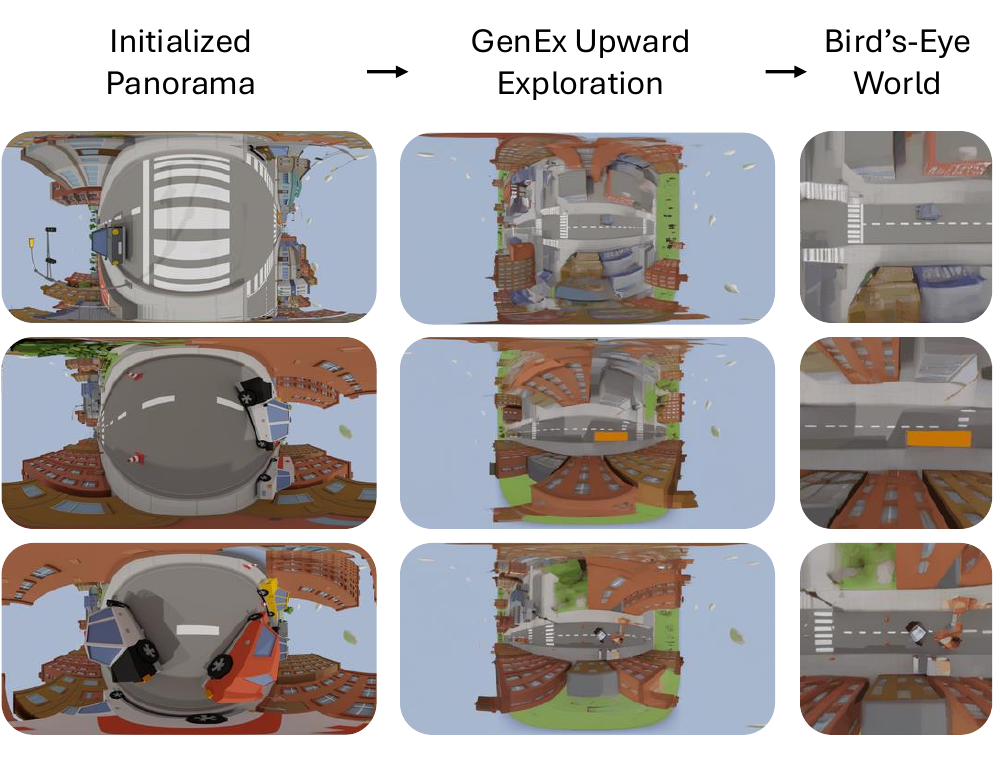}
\caption{\small Through generative exploration in \textit{z-axis}, we are able to generate the 2D bird-eye world view of the current scene.
}
\vspace{-5mm}
\label{fig:BEV}
\end{figure}

By exploring upward along the z-axis, our method generates top-down (bird’s-eye view) maps directly from a single panoramic image. As shown in \autoref{fig:BEV}, these overhead layouts give the agent an objective, third-person understanding of the scene, thereby improving reasoning.

\subsection{3D Consistency}
\begin{figure}[ht!]
\centering
    \includegraphics[width=\linewidth]{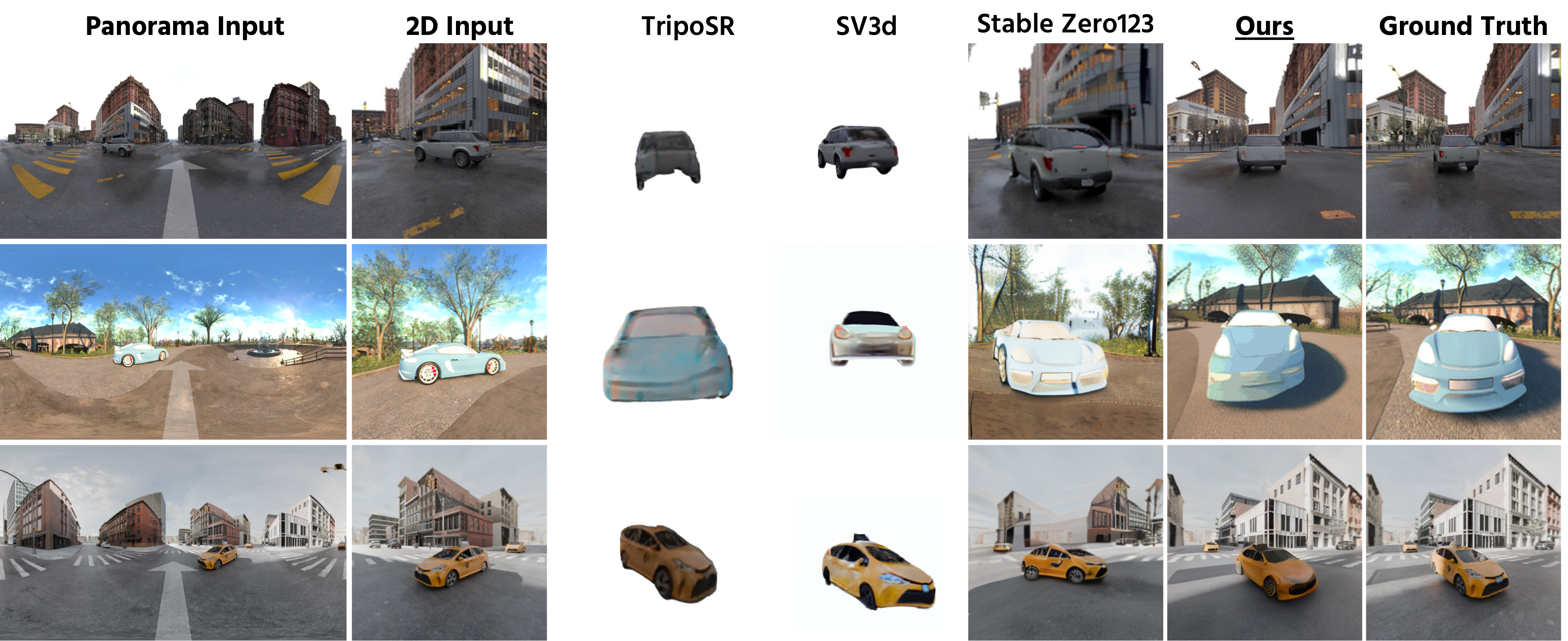}
    \caption{\small Through exploration, our model achieves higher quality in novel view synthesis for objects and better consistency in background synthesis, compared to SOTA 3D reconstruction models~\citep{voleti2024sv3d, tochilkin2024triposr, stablezero123}.
    }
    \label{fig:novel_view_image}
\end{figure}

Our method enables the generation of multi-view videos of an object through imaginative exploration with a path circling around it. Our model demonstrates superior performance compared with the SOTA open-source models. Importantly, it maintains near-perfect background consistency and effectively simulates scene lighting, object orientation, and 3D relationships as in \autoref{fig:novel_view_image}. 

\subsection{Active 3D Mapping in Generated Worlds}
\begin{figure}[ht!]
\begin{center}
\includegraphics[width=1.0\linewidth]{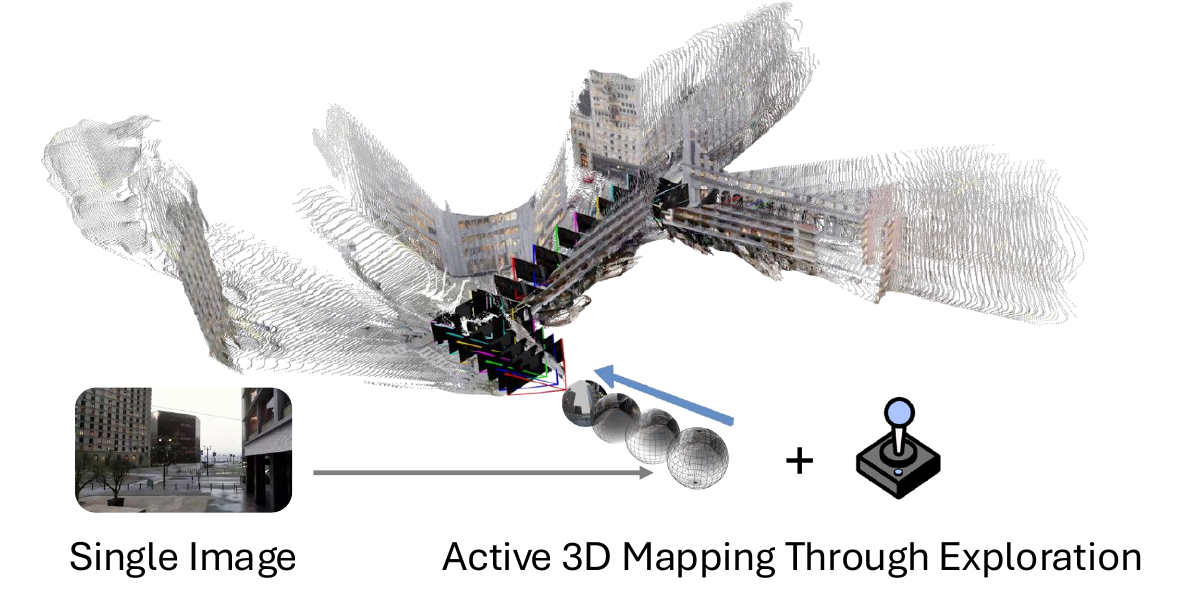}
\end{center}
\vspace{-0.2cm}
\caption{\small Active 3D mapping from a single image.
}
\label{fig:3d_mapping}
\end{figure}

When the agent actively explores the generative world, it continuously gathers observations that can be leveraged to reconstruct a 3D map using DUSt3R~\citep{wang2024dust3r}, shown in \autoref{fig:3d_mapping}.

\subsection{Embodied Decision Making}
We next evaluate the Imagination-Augmented Policy proposed in \S\ref{sec:eai} and share two key findings.

\textbf{Evaluation}.
We evaluate our Imagination-Augmented Policy (\S\ref{sec:imaginative_policy}) in \autoref{tab:imaginative_policy}. We extend the Genex-EQA in~\citep{lu2024generative} with a controlled counterpart for each scenario.
We use \textit{Unimodal} to refer to agents receiving only text context, while \textit{Multimodal} reasoning demonstrates LLM decision when prompted along with an egocentric visual view. GenEx shows the performance of models equipped as agents with a generative world explorer. We evaluate our Multi-Agent Imagination-Augmented Policy (\S\ref{sec:multiagent_policy}) in \autoref{tab:multiagent_policy}.

\begin{table}[ht!]
\centering
\renewcommand{\arraystretch}{1.5}
\setlength{\tabcolsep}{3pt}
\resizebox{0.47\textwidth}{!}{
\begin{tabular}{lccc}
\toprule
Method & Acc. (\%) & Confidence (\%) & Logic Acc. (\%) \\
\midrule\midrule
Random              & 25.00 & 25.00 & -     \\
% \cmidrule(l{0.1em}r{0.3em}){1-4} 
\midrule
Human Text-only     & 44.82 & 52.19 & 46.82 \\
Human with Image    & 91.50 & 80.22 & 70.93 \\
\human{Human with \textbf{GenEx}} & \human{\textbf{94.00}} & \human{\textbf{90.77}} & \human{\textbf{86.19}} \\
% \cmidrule(l{0.1em}r{0.3em}){1-4} 
\midrule\midrule
Unimodal Gemini-1.5 & 30.56 & 29.46 & 13.89 \\
Unimodal GPT-4o     & 27.71 & 26.38 & 20.22 \\
\cmidrule(l{0.1em}r{0.3em}){1-4} 
Multimodal Gemini-1.5 & 46.73 & 36.70 & 0.0  \\
Multimodal GPT-4o   & 46.10 & 44.10 & 12.51 \\
\cmidrule(l{0.1em}r{0.3em}){1-4} 
\gpt{\textbf{GPT4-o with GenEx}} & \gpt{\textbf{85.22}} & \gpt{\textbf{77.68}} & \gpt{\textbf{83.88}} \\
\bottomrule
\end{tabular}
}
\caption{Eval of Imagination-Augmented Policy. 
}
\label{tab:imaginative_policy}
\end{table}

\begin{table}[ht!]
\centering
\vspace{-1mm}
\renewcommand{\arraystretch}{1.5}
\setlength{\tabcolsep}{3pt}
\resizebox{0.47\textwidth}{!}{
\begin{tabular}{lccc}
\toprule
Method & Acc. (\%) & Confidence (\%) & Logic Acc. (\%) \\
\midrule\midrule
Random              & 25.00 & 25.00 & -    \\
% \cmidrule(l{0.1em}r{0.3em}){1-4} 
\midrule
Human Text-only     & 21.21 & 11.56 & 13.50 \\
Human with Image    & 55.24 & 58.67 & 46.49 \\
\human{Human with \textbf{GenEx}} & \human{\textbf{77.41}} & \human{\textbf{71.54}} & \human{\textbf{72.73}} \\
% \cmidrule(l{0.1em}r{0.3em}){1-4} 
\midrule\midrule
Unimodal Gemini-1.5 & 26.04 & 24.37 & 5.56  \\
Unimodal GPT-4o     & 25.88 & 26.99 & 5.00  \\
\cmidrule(l{0.1em}r{0.3em}){1-4}  
Multimodal Gemini-1.5 & 11.54 & 15.35 & 0.0  \\
Multimodal GPT-4o   & 21.88 & 21.16 & 6.25 \\
\cmidrule(l{0.1em}r{0.3em}){1-4} 
\gpt{\textbf{GPT4-o with GenEx}} & \gpt{\textbf{94.87}} & \gpt{\textbf{69.21}} & \gpt{\textbf{72.11}} \\
\bottomrule
\end{tabular}
}
\caption{Evaluation of Multi-Agent Imagination-Augmented Policy.}
\vspace{-2mm}
\label{tab:multiagent_policy}
\end{table}

\vfill\eject
\textbf{Findings}. We identified two findings based on the results from human policy (\colorbox{humancolor}{grey row}) and GenEx-enhanced GPT policy (\colorbox{transparentblue}{blue row}).
\begin{itemize}[leftmargin=*,itemsep=0pt, before=\vspace{-0.6\baselineskip}, after=\vspace{-\baselineskip}]

\item \textit{Vision without imagination can be misleading for GPTs}.
Interestingly, a unimodal response that relies solely on the environment's text description often outperforms its multimodal counterparts, which incorporate both text and egocentric visual inputs. This suggests that vision without imagination can be misleading, as it may lead to incorrect inferences due to the lack of spatial context and relying only on language-based commonsense reasoning.
This highlights the importance of integrating visual imagination to enhance the accuracy and reliability of the agent's decision-making processes.
\vspace{1mm}
\item \textit{GenEx has the potential to enhance cognitive abilities for humans.} Human performance results reveal several key insights. First, individuals using both visual and textual information achieve significantly higher decision accuracy compared to those relying solely on text. This indicates that multimodal inputs enhance reasoning.
Secondly, when provided with imagined videos generated by GenEx, humans make even more accurate and informed decisions than in the conventional image-only setting, especially in multi-agent scenarios that require advanced spatial reasoning. These findings demonstrate GenEx's potential to enhance cognitive abilities for effective social collaboration and situational awareness.
\end{itemize}

\clearpage
\section{6. Discussion}
\textbf{Related works}.
Advances in single-image 3D modeling~\citep{tewari2023diffusion, yu2024wonderworld} enable novel view synthesis but are limited by render distances or fields of view, relying heavily on depth estimator. Meanwhile, video generation methods~\citep{sora, blattmann2023svd, kondratyuk2023videopoet} excel at producing diverse videos but often lack physical grounding, reducing their utility for exploration. Video generation models~\citep{du2023video, yang2024video, wang2024language, bu2024closed, du2024learning} are capable of directly synthesizing visual plans for decision-making, but world exploration for imagined observations remains unexamined. Our approach unites these domains by drawing on physically grounded data to generate 3D-consistent, explorable worlds and advance embodied AI.

\textbf{Extension to our earlier work}. Our earlier work~\citep{lu2024generative}, published on arXiv in November 2024, conceptualized world transitions, exploration, and applications in embodied AI, but it did not address the crucial aspect of world initialization from a single image.

\textbf{Relation to concurrent industrial progress}. 
WorldLabs~\citep{worldlabs} recently released demos of anime-world generation from a single image. 
DeepMind~\citep{genie2} released a blog on interactive world models.
Our work complements these ongoing industrial efforts, jointly contributing toward a shared vision: creating rich, interactive, 3D-consistent generative worlds. Importantly, we offer our technical details. Beyond this, we also introduce the concept of an Imagination-Augmented Policy by exploring the generative world, further expanding the frontiers of embodied AI.

\textbf{Challenges}. Bridging imaginative and real-world environments remains a core challenge in AI. Current approaches rely on physical engines. Future work must address several key limitations, including sim-to-real adaptation, real sensor integration, dynamic conditions, and ethical safeguards, to ultimately enable reliable deployment of embodied AI in diverse physical settings.
\vfill\eject

\section{7. Conclusion}
We introduce \textbf{GenEx}, a platform that \textbf{Gen}erates an \textbf{Ex}plorable world and enables agents, either instructed by human users or a GPT, to freely explore in this imaginative panoramic world.
By generating 3D-consistent environments from a single image, our approach enables the creation of immersive and interactive worlds offering a boundless landscape, grounded in the physical world, and explored by agents.
We demonstrate diverse applications of GenEx, showing that this generative explorable world technique can create diverse and consistent 3D environments, build active 3D mappings, and advance embodied decision-making by allowing agents to create more informed and effective plans. Furthermore, GenEx’s framework supports multi-agent interactions, paving the way for more advanced and cooperative AI systems. This work marks an advancement toward real-world navigation, interactive gaming, and achieving human-like intelligence in embodied AI.

\section*{Author Contributions}
We list author contributions here alphabetically by last name. Please direct all correspondence to the project lead \textbf{\href{https://beckschen.github.io/}{\textcolor{dmred500}{Jieneng Chen}}} (\href{mailto:jchen293@jhu.edu}{\textcolor{dmred500}{jchen293@jh.edu}}). 

\subsection*{Core Contributors}
\begin{itemize}
     \item \textbf{\href{https://taiminglu.com/}{\textcolor{dmred500}{Taiming Lu}}}: project leadership, data engine, model research and pipeline, infrastructure
     \item \textbf{\href{https://www.tshu.io/}{\textcolor{dmred500}{Tianmin Shu}}}: embodied policy research, writing, revising, technical advice
     \item \textbf{\href{https://lambert-x.github.io/}{\textcolor{dmred500}{Junfei Xiao}}}: image-to-panorama data and model research, writing, editing
\end{itemize}
\vspace{-6mm}
\subsection*{Contributors and Advisors}
\vspace{-3mm}
\begin{itemize}
    \item \textbf{\href{https://engineering.jhu.edu/faculty/rama-chellappa/}{\textcolor{dmred500}{Rama Chellappa}}}: device support, advice
    \item \textbf{\href{https://danielkhashabi.com/}{\textcolor{dmred500}{Daniel Khashabi}}}: writing, technical advice
    \item \textbf{\href{https://sites.google.com/view/cheng-peng/home}{\textcolor{dmred500}{Cheng Peng}}}: data support, editing
      \item \textbf{\href{https://jiahaoplus.github.io/}{\textcolor{dmred500}{Jiahao Wang}}}: math, postprocessing, editing
    \item \textbf{\href{https://weichen582.github.io/}{\textcolor{dmred500}{Chen Wei}}}: revising, editing, writing advice
 \item \textbf{\href{https://openreview.net/profile?id=~Luoxin_Ye1}{\textcolor{dmred500}{Luoxin Ye}}}:  model, postprocessing
\item \textbf{\href{https://cogsci.jhu.edu/directory/alan-yuille/}{\textcolor{dmred500}{Alan L. Yuille}}}: math revising, funding, editing, writing advice, technical advice
\end{itemize}
% \vfill\eject

\clearpage
{
\small
\bibliographystyle{abbrvnat}
\nobibliography*
\bibliography{refs}
}
\newpage
\appendix
\onecolumn
\section{Appendix}
\subsection{Preliminary: Equirectangular Panorama Images}
\label{sec:panorama_explained}
\begin{figure}[ht!]
\begin{center}
\includegraphics[width=\linewidth]{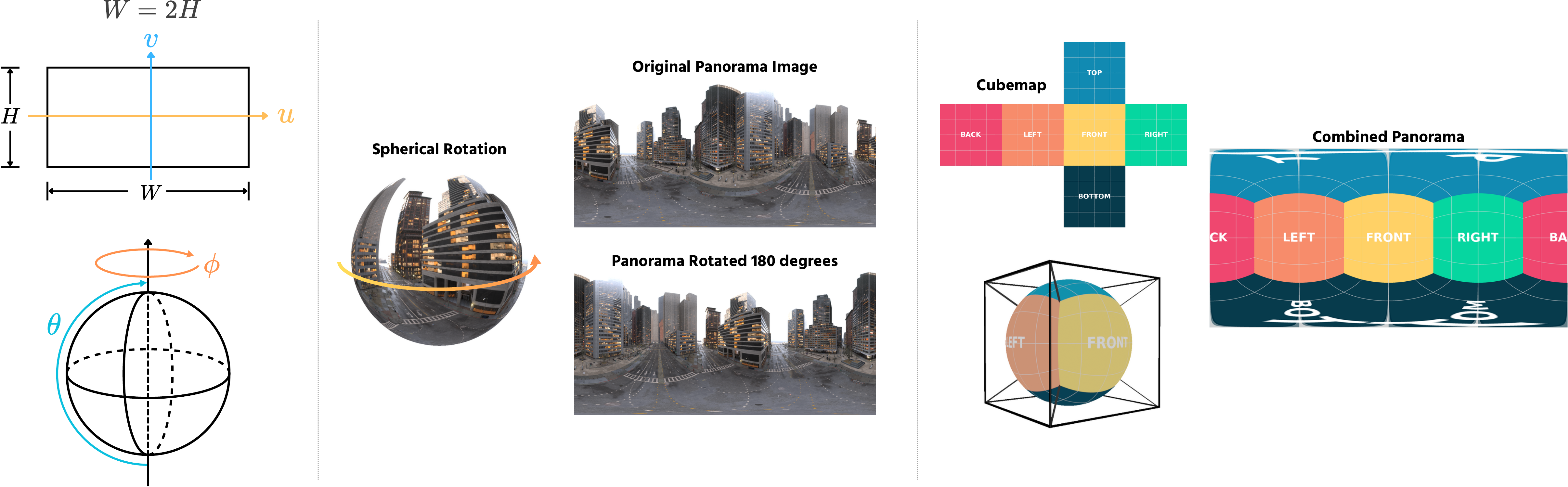}
\end{center}
\caption{Left: Pixel Grid coordinate and Spherical Polar coordinate systems; Middle: rotation in Spherical coordinates corresponds to a rotation in the 2D image; Right: expansion from panorama to cubemap or composition in reverse.}
\label{fig:sup_panorama_demo}
\end{figure}

\subsubsection{Coordinate Systems}
An \textit{Equirectangular Panorama Image} captures all perspectives from an egocentric viewpoint into a 2D image. Essentially, it represents a spherical coordinate system on a 2D grid.
\begin{definition}[Spherical polar coordinate system]\label{def:spherical_coor}
$\mathcal{S}$: Taking the origin as the central point, a point in this system is represented by coordinates $(\phi, \theta, r) \in \mathcal{S}$, where $\phi$ denotes the longitude, $\theta$ the latitude, and $r$ the radial distance from the origin. The ranges for these coordinates are $\phi \in [-\pi, \pi)$, $\theta \in [-\pi/2, \pi/2]$, and $r > 0$.
\end{definition}

\begin{definition}[Cartesian coordinate system for panoramic image]\label{def:cartesian_coor}
$\mathcal{P}$: In this system, a pixel is identified by the coordinates $(u,v) \in \mathcal{P}$, where $u$ and $v$ correspond to the column and row positions on the 2D panoramic image plane, respectively. Here, $u$ ranges from $0$ to $W-1$ and $v$ ranges from $0$ to $H-1$.
\end{definition}
\begin{definition}[Sphere-to-Cartesian Coordinate Transformation]\label{def:coor_transform} The transformation between the spherical polar coordinates and the panoramic pixel grid coordinates can be defined by the following functions:
    \begin{align}
        f_{\mathcal{S} \to \mathcal{P}}(\phi, \theta) &= \left( \frac{W}{2\pi} (\phi + \pi), \frac{H}{\pi} \left(\frac{\pi}{2} - \theta\right) \right) \label{eq:s_to_p} \\
        f_{\mathcal{P} \to \mathcal{S}}(u, v) &= \left( \frac{2\pi u}{W} - \pi, \frac{\pi}{2} - \frac{\pi v}{H} \right) \label{eq:p_to_s}
    \end{align}
    Here, the function $f_{\mathcal{S} \to \mathcal{P}}$ maps the spherical coordinates $(\phi, \theta)$ to the pixel coordinates $(u,v)$, and the inverse function $f_{\mathcal{P} \to \mathcal{S}}$ maps the pixel coordinates $(u,v)$ back to the spherical coordinates $(\phi, \theta)$. This transformation ensures that the entire spherical surface is represented on the 2D panoramic image.

\end{definition}
% Here, the longitude $\phi$ is mapped to the horizontal coordinate $u$, and the latitude $\theta$ is mapped to the vertical coordinate $v$ on the 2D grid. This transformation ensures that the entire spherical surface is represented on the 2D panoramic image.

Panorama effectively stores every perspective of the world from a single location. In our work, due to the nature of panoramic images, we are able to preserve the global context during spatial navigation. This allows us to maintain consistency in world information from the conditional image, ensuring that the generated content aligns coherently with the surrounding environment.
    
\subsubsection{Panorama Image transformations}
The spherical format allows various image processing tasks. For example, the image can be rotated by an arbitrary angle without any loss of information due to the spherical representation. Additionally, it can be broken down into cubemaps for 2D visualization, as shown in \autoref{fig:sup_panorama_demo}.
\begin{definition}[Rotation Transformation in Spherical Polar Coordinate System]\label{def:rotate_transform}  Since a panorama image is in a spherical format, we can rotate the image to face a different angle while preserving the original image quality. The rotation can be performed using the following formula:

\begin{equation}
\mathcal{T}
(u, v, \Delta\phi, \Delta\theta) = f_{\mathcal{S} \to \mathcal{P}}\left(\mathcal{R}\left(f_{\mathcal{P} \to \mathcal{S}}(u, v), \Delta\phi, \Delta\theta\right)\right)
\label{eq:composite_rotation_T_appendix}
\end{equation}

Where the rotation function \(\mathcal{R}\) is defined as:

\begin{equation}
\mathcal{R}(\phi, \theta, \Delta\phi, \Delta\theta) = \left(\phi + \Delta\phi \ (\text{mod } 2\pi), \theta + \Delta\theta \ (\text{mod } \pi)\right)
\label{eq:spherical_rotation}
\end{equation}
\end{definition}

If there is no explicit input, both $\Delta\phi$ and $ \Delta\theta$ can be set to 0.

\end{document}